\documentclass[10pt,twocolumn,letterpaper]{article}

\usepackage{iccv}
\usepackage{times}
\usepackage{epsfig}
\usepackage{graphicx}
\usepackage{amsmath}
\usepackage{amssymb}

\usepackage{booktabs}
\usepackage{multirow}
\usepackage[dvipsnames]{xcolor}
\usepackage{colortbl}
\usepackage{bm}

\usepackage[pagebackref=true,breaklinks=true,letterpaper=true,colorlinks,bookmarks=false]{hyperref}

\iccvfinalcopy 

\def\iccvPaperID{2543} 

\ificcvfinal\fi

\DeclareMathOperator*{\argmax}{arg\,max}

\def\method{SIO}

\def\upcolor{LimeGreen}
\def\downcolor{Red}
\definecolor{LightCyan}{rgb}{0.8,1,1}
\newcolumntype{b}{>{\columncolor{gray!10}}c}

\newcommand{\Din}{\mathcal{D}_{\text{in}}}
\newcommand{\Dinr}{\mathcal{D}_{\text{in}}^{\text{real}}}
\newcommand{\Dins}{\mathcal{D}_{\text{in}}^{\text{syn}}}

\newcommand{\Dout}{\mathcal{D}_{\text{out}}}
\newcommand{\DoutOE}{\mathcal{D}_{\text{out}}^{\text{OE}}}

\begin{document}

\title{SIO: \underline{S}ynthetic \underline{I}n-Distribution Data Benefits \underline{O}ut-of-Distribution Detection}


\author{Jingyang Zhang\textsuperscript{\textdagger}, Nathan Inkawhich\textsuperscript{*}, Randolph Linderman\textsuperscript{\textdagger},
Ryan Luley\textsuperscript{*},
Yiran Chen\textsuperscript{\textdagger}, Hai Li\textsuperscript{\textdagger}\\
\textsuperscript{\textdagger}Duke University, 
\textsuperscript{*}Air Force Research Laboratory\\
{\tt\small jingyang.zhang@duke.edu}}

\maketitle
\ificcvfinal\thispagestyle{empty}\fi


\begin{abstract}
    Building up reliable Out-of-Distribution (OOD) detectors is challenging, often requiring the use of OOD data during training.
    In this work, we develop a data-driven approach which is distinct and complementary to existing works: Instead of using external OOD data, we fully exploit the internal in-distribution (ID) training set by utilizing generative models to produce additional synthetic ID images.
    The classifier is then trained using a novel objective that computes weighted loss on real and synthetic ID samples together.
    Our training framework, which is termed SIO, serves as a ``plug-and-play'' technique that is designed to be compatible with existing and future OOD detection algorithms, including the ones that leverage available OOD training data.
    Our experiments on CIFAR-10, CIFAR-100, and ImageNet variants demonstrate that SIO consistently improves the performance of nearly all state-of-the-art (SOTA) OOD detection algorithms.
    For instance, on the challenging CIFAR-10 v.s. CIFAR-100 detection problem, SIO improves the average OOD detection AUROC of 18 existing methods from 86.25\% to 89.04\% and achieves a new SOTA of 92.94\% according to the OpenOOD benchmark.
    Code is available at \url{https://github.com/zjysteven/SIO}.
\end{abstract}

\section{Introduction}
\label{sec:intro}

Being able to identify unknowns is of the utmost importance for intelligent systems to reliably operate in open world settings.
In the domain of image classification, this challenge is known as \textit{Out-of-Distribution (OOD) Detection}. 
The goal of OOD detection is to enable the classifier to identify samples that do not belong to one of the known, in-distribution (ID) categories during inference.
However, OOD detection has long been a difficult task due to the implicit closed-world assumption adopted by standard neural network training.
For instance, without certain efforts, a basic CIFAR-10 classifier will confidently predict SVHN digits as one of its classes \cite{whyrelu19cvpr}.

Recent years have seen plenty of OOD detection works, which we roughly divide into three categories.
1) \textit{Inference techniques} study post-hoc scoring rules that best separate ID and OOD data with a pre-trained model(the score indicates each sample's ``OOD-ness'') \cite{openmax16cvpr,msp17iclr,guo2017calibration,odin18iclr,energyood20nips,gradnorm21neurips,react21neurips,species22icml,species22icml,haoqi2022vim,sun2022knnood,sun2021dice}.
2) \textit{Specialized training algorithms} induce more suitable structure \cite{godin20cvpr,wei2022mitigating,cutmix19cvpr} or feature distribution \cite{vos22iclr} inside the model via training to allow for better OOD detection at inference time.
3) \textit{Data-driven methods} utilize additional input data for improvements.
Particularly, existing works incorporate external OOD samples to either let the model learn OOD detection in a supervised way \cite{oe18nips} or mix them with ID images as a data augmentation \cite{hendrycks2021pixmix,mixoe}.
Not surprisingly, data-driven ones are among the most effective methods as they explicitly bring in additional information outside the ID space.

\begin{figure}[t]
\centering
   \includegraphics[width=0.95\linewidth]{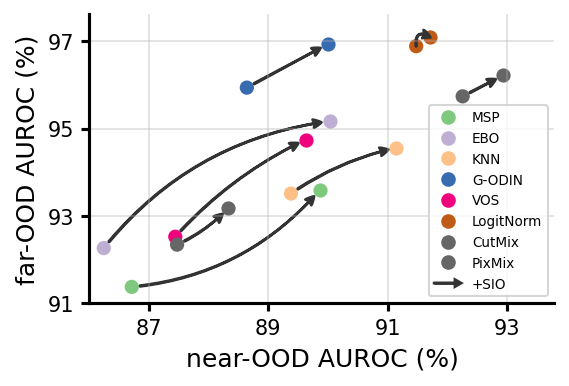}
   \vspace{-1mm}
   \caption{Results on CIFAR-10. SIO is able to yield improvements on top of multiple methods against both near- and far-OOD. See full results in Section \ref{sec:4.1} Table \ref{tab:cifar}.}
   \vspace{-3mm}
\label{fig:intro}
\end{figure}

In this work, we seek to further improve OOD detection by following along the data-driven path.
However, we think in an entirely different and orthogonal direction to existing approaches: While external OOD data could be helpful, \textit{we wonder if the original, internal ID data can be better exploited to benefit OOD detection.}
To answer this question, we propose using generative models (\eg, GANs \cite{stylegan2ada,biggan}, diffusion models \cite{ddpm}) to capture the ID data distribution and to produce additional synthetic ID images to augment the existing real ID set.

On the face of it, one may think that synthetic ID images would not benefit OOD detection and could even have a negative impact for two reasons.
First, since synthetic ID images do not provide any explicit information about the OOD samples, they may not seem useful for improving OOD detection.
Second, prior studies \cite{cas} have shown that naively adding synthetic images into the real training set can lower classification accuracy, which in turn may lead to inferior OOD detection performance according to the correlation between OOD detection rate and ID accuracy \cite{vaze2021open}.

Contrary to initial assumptions, we reveal that under a reasonable weighting scheme, synthetic ID images can indeed be advantageous when integrated with real ID images. 
Concretely, we propose a novel and generic training framework which employs a weighted sum of the loss computed on both real and synthetic ID images to train the classifier. 
We term this framework SIO (\underline{S}ynthetic \underline{I}D data for \underline{O}OD detection).
The proposed SIO framework is distinct and complementary to existing OOD detection methodologies.
It can be seen as a ``plug-and-play'' technique that can be easily integrated into most (if not all) dedicated OOD detection approaches, including those that incorporate OOD samples in the training phase (see Section \ref{sec:3.2} for detailed discussion).
Practically, we design a lightweight implementation of SIO that only modifies the ID batch sampling. 
This guarantees that SIO does not introduce any computational overhead compared to real data-only training, facilitating efficient use and fair comparison.

We conduct a thorough evaluation of SIO in combination with a variety of state-of-the-art OOD detection methods on two widely-used image classification datasets, CIFAR-10 and CIFAR-100, using the OpenOOD benchmark \cite{yang2022openood}. 
Our results indicate that SIO consistently improves OOD detection performance on top of multiple advanced OOD methods (see Figure \ref{fig:intro} for some of the results on CIFAR-10). 
Notably, by applying SIO we achieve new state-of-the-art results in 3 out of the 4 evaluation settings within the OpenOOD benchmark.
To demonstrate scalability to high-resolution images, we further experiment with two ImageNet variants. 
Additionally, we conduct extensive analyses to evaluate the robustness of SIO against hyperparameters, such as the choice of generative models.
Finally, we observe that SIO improves OOD detection rates even if it does not improve ID classification accuracy, providing a more comprehensive view of the OOD detection rate vs. ID accuracy correlation \cite{vaze2021open}.

\section{Background}
\label{sec:background}

\subsection{Problem statement}

In this work we consider OOD detection in the context of multi-class image classification, where the goals are 1) training a base classifier that can accurately classify ID data and 2) building an OOD detector on top of the base classifier that accurately distinguishes OOD from ID samples.
We now formulate this problem to facilitate the discussion.

\noindent \textbf{Training.}
There are in general two types of training schemes adopted by existing works.
The first one trains the base classifier $f$ only on the labeled (real) ID training data sampled from the in-distribution set $\Din$:
\begin{equation}
    \min_{f}\mathbb{E}_{(\bm{x},y)\sim \Din}L(\bm{x},y;f).
\label{eq:id_only_training}
\end{equation}
Here, Equation \ref{eq:id_only_training} is a high-level abstraction of the true optimization objectives used in practice.
For instance, $L$ could be as simple as the standard cross-entropy loss, \ie, $L(\bm{x},y;f)=H(y,\sigma(f(\bm{x})))$, where $f(\bm{x})$ is the logit vector, $\sigma$ is the softmax function, and $y$ is the one-hot representation of the ground-truth label.
$L$ could also involve additional regularization as in \cite{vos22iclr,wei2022mitigating}.

The other type of training assumes the availability of an external set of unlabeled OOD/outlier samples $\Dout$ and trains the classifier with ID and OOD samples together:
\begin{equation}
    \min_{f}\mathbb{E}_{(\bm{x},y)\sim\Din,\hat{\bm{x}}\sim\Dout}L(\bm{x},y,\hat{\bm{x}};f).
\label{eq:id_ood_training}
\end{equation}
For example, the loss function of Outlier Exposure (OE) \cite{oe18nips} is $L(\bm{x},\hat{\bm{x}},y;f)=H(y,\sigma(f(\bm{x}))+\lambda\cdot H(\mathcal{U},\sigma(f(\hat{\bm{x}}))$, where $\mathcal{U}$ is the uniform distribution across the known categories, and $\lambda$ is a weighting term.
Later in Section \ref{sec:3.2}, we will demonstrate the compatibility of our proposed SIO with both of these two training schemes.

\noindent \textbf{Inference.}
After training, the OOD detector $g$ is built upon the base classifier $f$ with a scoring module $s$:
\begin{equation}
    g(\bm{x};f,\tau)= 
    \begin{cases}
        \text{OOD,} & \text{if}~ s(\bm{x};f) \geq \tau\\
        \text{ID,}              & \text{otherwise}
    \end{cases}.
\label{eq:detector}
\end{equation}
The scoring module $s$ will assign a score to each sample which indicates its ``OOD-ness''.
Again, $s$ is a high-level representation of the scoring mechanism.
Examples of $s$ include $s(\bm{x};f)=\max_i\sigma(f(\bm{x})_i)$ (MSP \cite{msp17iclr}) and $s(\bm{x};f)=\sum_{i=1}^K e^{f(\bm{x})_i}$ (EBO \cite{energyood20nips}; $K$ is the number of ID categories).
Here $\tau$ is an application-dependent threshold.
The detector $g$ is used to determine whether each incoming sample is ID or OOD.
For OOD samples, the base classifier will refrain from making any predictions.

\subsection{Related works}

\noindent \textbf{1) OOD detection methodologies.}
There have been many works on OOD detection since it emerged as a research problem.
We refer readers to \cite{survey} for a comprehensive survey, while we focus on several top-performing methods that we consider in this work.
As aforementioned, we roughly categorize existing works into three groups.

The first line of works focus on the design of the post-hoc scoring rule $s$ (Equation \ref{eq:detector}), while assuming that the base classifier is pre-trained (usually with the standard cross-entropy loss).
In general, the outputs from the model's decision space (\eg, the final linear layer or the penultimate layer) could derive informative scores.
For example, MSP \cite{msp17iclr} and MLS \cite{species22icml} directly use the (negative) maximum softmax probability and maximum logit value as the score, respectively.
Other works post-process the network's outputs to enlarge the difference between ID and OOD.
Examples include softening softmax probabilities with temperature scaling \cite{guo2017calibration}, applying energy function to the logits \cite{energyood20nips}, rectifying activations with thresholding \cite{react21neurips}, and applying KNN to the penultimate layer's features \cite{sun2022knnood}, etc.

While some researchers focus on inference techniques, others investigate specialized training algorithms that involve developing more advanced $L$ in Equation \ref{eq:id_only_training}.
Notably, G-ODIN \cite{godin20cvpr} trains a dividend/divisor structure to decompose the softmax confidence.
VOS \cite{vos22iclr} encourages the learned representations of the classifier to shape like a mixture of Gaussian distributions.
LogitNorm \cite{wei2022mitigating} trains the model with normalized logit vectors as the unit sphere provides more discriminative information for distinguishing ID and OOD samples.
Meanwhile, using certain tricks such as longer training, stronger data augmentation, and more complex learning rate schedule has also been shown to benefit OOD detection \cite{vaze2021open}.

The final group of works proposes data-driven approaches to improve OOD detection.
In particular, OE \cite{oe18nips} collects real OOD samples and explicitly lets the model learn OOD detection in a supervised manner.
PixMix \cite{hendrycks2021pixmix} applies pixel-level mixing operations between ID images and low-level OOD images (which exhibit certain visual patterns but do not have clear semantics) as an augmentation.
Additionally, pre-training with a huge labeled dataset has been shown to be beneficial for OOD detection in downstream tasks (\eg, using ImageNet pre-training for CIFAR-10) \cite{hendrycks2019using}.
\textit{It is noteworthy that all these data-driven methods require external data beyond the original ID training set.}

While SIO, like other data-driven approaches, introduces additional input images, it stands out from existing methods by utilizing synthetic ID images instead of external OOD images.
This approach is particularly useful in data-scarce scenarios, where external data may not be readily available.
Furthermore, SIO is inherently complementary to all previously discussed methods, including data-driven ones. 
We provide detailed discussions and empirical evidence to support this claim in the subsequent sections.


\noindent \textbf{2) Synthetic images for OOD detection.}
Although previous studies in OOD detection research have used synthetic images produced by generative models, they have all focused on synthesizing OOD images \cite{gen-openmax,counterfactual,confcal18iclr}.
However, synthesizing OOD images is a challenging task as there could be a lack of real OOD data for supervision, and the distribution of the open space is too broad to capture. 
As a result, previous attempts have not yielded superior performance, with models trained on synthetic OOD images underperforming those trained on real OOD data \cite{oe18nips} or even simple ID-only training baselines \cite{vaze2021open}.


In sharp contrast, what we propose is using generative models to synthesize ID images, which is much easier a task since the target distribution is well-defined (characterized by the ID training set).
In addition, we demonstrate that incorporating synthetic ID samples can significantly improve the performance of multiple OOD detection methods.

\begin{figure}[t]
\centering
   \includegraphics[width=0.95\linewidth]{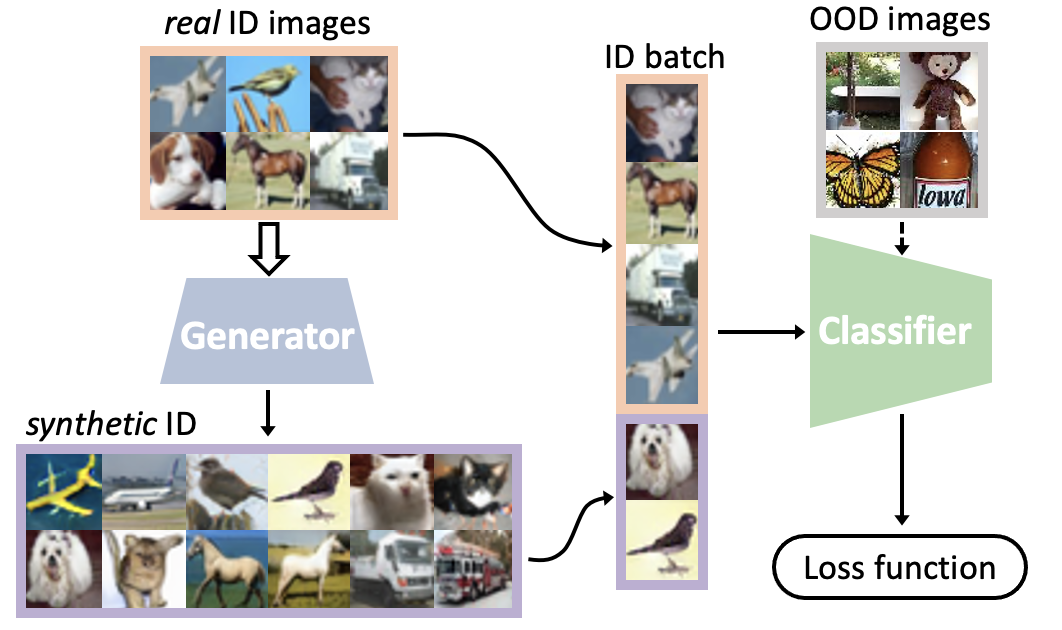}
   \caption{Overview of the proposed SIO framework. First, a large number of synthetic ID samples are generated (offline) using a generative model that is trained with real ID images. Then, we combine the real and synthetic ID samples to train the classifier in a weighted fashion described in Equation \ref{eq:SIO_id_only_training}. The specific loss function can be selected from any existing OOD training methods. External OOD samples can also be incorporated into the training depending on their availability (Equation \ref{eq:SIO_id_ood_training}).
   }
   \vspace{-2mm}
\label{fig:method}
\end{figure}

\noindent \textbf{3) Synthetic samples for other performance measures.}
Expanding the real training set with synthetic samples has been demonstrated to help with adversarial training \cite{gowal2021improving}, where overfitting is the primary challenge that hinders better adversarial robustness.
In standard, non-adversarial settings, however, including synthetic samples has not yet led to improvements in, \eg, classification accuracy \cite{cas}.
To the best of our knowledge, we are the first to investigate and demonstrate that synthetic ID images can enhance OOD detection.
Our study provides a new avenue for exploring the use of synthetic data in other areas of machine learning.


\section{Methodology}

We now describe the proposed SIO as a two-step procedure, including several technical details.
See Figure \ref{fig:method} for an overview.

\subsection{Generating synthetic ID training set $\Dins$}
\label{sec:3.1}

Given the real ID training set $\Dinr$, we train a generative model to capture the ID data distribution, which can then be used to generate a large number of synthetic ID images by sampling from the synthetic distribution $\Dins$.
The training step can be skipped if pre-trained generative models are available. For instance, we utilize off-the-shelf generative models that are pre-trained on standard ID datasets (\eg, CIFAR-10) in most of our experiments.
It is worth noting that a generator trained on a different (potentially larger) dataset can also be used, as long as it knows the concept of ID categories (\eg, an ImageNet generator can be employed for a Tiny ImageNet classifier).

The way the labels of synthetic samples are obtained can vary depending on the case. 
If the real ID dataset $\Dinr$ has labels (which is typically the case), the generative model can be trained in a class-conditional way, and we can directly sample labeled synthetic data $(\tilde{\bm{x}},\tilde{y})\sim\Dins$. 
If $\Dinr$ is unlabeled (\eg, in a self-supervised setting) or the generative model is unconditional, we can only sample unlabeled synthetic data $\tilde{\bm{x}}\sim\Dins$. 
In this case, we use a classifier $f$ pre-trained on the real data to produce pseudo-labels, i.e., $\tilde{y}=\argmax_if(\tilde{\bm{x}})_i$. 
By default, we assume that the generative model is conditional unless otherwise stated.

In this work, we consider the generation step as offline, meaning that we pre-generate a fixed number of synthetic images.
Therefore, it does not add any complexity to the online training step. 
However, if there are sufficient computing resources, synthetic samples can certainly be generated on the fly. In fact, our experiments indicate that more synthetic samples generally lead to better performance.

\begin{table*}[!t]
\centering
\caption{Results in terms of detection AUROC (\%) on CIFAR-10/100 datasets. All numbers are percentages and are averaged over 3 runs. In most cases introducing synthetic ID data leads to \textcolor{\upcolor}{improvements}. Note that these numbers cannot be directly compared with those reported in the OpenOOD paper \cite{yang2022openood} due to differences in training and evaluation setup. See text for detailed description.}
\label{tab:cifar}
\vspace{2mm}
\resizebox{0.95\textwidth}{!}{
{\setlength\doublerulesep{1.2pt}
\begin{tabular}{lcbcbcbcbcbcb}
\toprule[1.5pt]\midrule[0.8pt]
\multirow{3}{*}{Method} &\multicolumn{6}{c}{CIFAR-10} &\multicolumn{6}{c}{CIFAR-100} \\ \cmidrule(lr){2-7} \cmidrule(lr){8-13}
&\multicolumn{2}{c}{near-OOD} 
&\multicolumn{2}{c}{far-OOD}
&\multicolumn{2}{c}{ID Accuracy}
&\multicolumn{2}{c}{near-OOD} 
&\multicolumn{2}{c}{far-OOD}
&\multicolumn{2}{c}{ID Accuracy}\\ 
\cmidrule(lr){2-3} \cmidrule(lr){4-5} \cmidrule(lr){6-7} \cmidrule(lr){8-9}
\cmidrule(lr){10-11} \cmidrule(lr){12-13}
& baseline & +\method & baseline & +\method & baseline & +\method & baseline & +\method & baseline & +\method & baseline & +\method \\
\midrule\midrule

\multicolumn{9}{l}{\textbf{- Inference techniques}} \vspace{.05cm} \\

OpenMax \cite{openmax16cvpr} &85.14 &\textcolor{\upcolor}{87.99} &90.08 &\textcolor{\upcolor}{92.50} & & &73.78 &\textcolor{\upcolor}{75.12} &70.33 &\textcolor{\upcolor}{73.56} & & \\
MSP \cite{msp17iclr} &86.71 &\textcolor{\upcolor}{89.86} &91.37 &\textcolor{\upcolor}{93.59} & & &79.56 &\textcolor{\downcolor}{79.04} &80.10 &\textcolor{\upcolor}{82.12} & & \\
TempScale \cite{guo2017calibration} &86.30 &\textcolor{\upcolor}{89.72} &92.39 &\textcolor{\upcolor}{94.57} & & &79.54 &\textcolor{\downcolor}{79.38} &83.93 &\textcolor{\upcolor}{84.71} & & \\
ODIN \cite{odin18iclr} &77.86 &\textcolor{\upcolor}{85.71} &86.68 &\textcolor{\upcolor}{92.62} & & &79.12 &\textcolor{\downcolor}{78.89} &81.02 &\textcolor{\upcolor}{84.23} & & \\
EBO \cite{energyood20nips} &86.24 &\textcolor{\upcolor}{90.03} &92.26 &\textcolor{\upcolor}{95.17} & & &80.17 &\textcolor{\upcolor}{80.20} &81.51 &\textcolor{\upcolor}{85.50} & & \\
ReAct \cite{react21neurips} &84.63 &\textcolor{\upcolor}{89.29} &90.86 &\textcolor{\upcolor}{94.40} & & &72.52 &\textcolor{\downcolor}{69.45} &78.33 &\textcolor{\upcolor}{85.84} & & \\
MLS \cite{species22icml} &86.17 &\textcolor{\upcolor}{89.99} &92.11 &\textcolor{\upcolor}{95.00} & & &80.22 &\textcolor{\upcolor}{80.34} &81.33 &\textcolor{\upcolor}{85.25} & & \\
KLM \cite{species22icml} &79.85 &\textcolor{\upcolor}{81.50} &85.41 &\textcolor{\upcolor}{87.57} & & &73.38 &\textcolor{\upcolor}{75.12} &76.17 &\textcolor{\upcolor}{80.32} & & \\
VIM \cite{haoqi2022vim} &84.16 &\textcolor{\upcolor}{88.23} &90.43 &\textcolor{\upcolor}{92.51} & & &63.70 &\textcolor{\upcolor}{72.00} &74.06 &\textcolor{\upcolor}{82.89} & & \\
KNN \cite{sun2022knnood} &89.38 &\textcolor{\upcolor}{91.15} &93.51 &\textcolor{\upcolor}{94.56} & & &77.75 &\textcolor{\downcolor}{77.65} &82.34 &\textcolor{\upcolor}{86.13} & & \\
DICE \cite{sun2021dice} &80.20 &\textcolor{\upcolor}{85.54} &87.78 &\textcolor{\upcolor}{92.56} &\multirow{-11}{*}{94.93} &\multirow{-11}{*}{\textcolor{\upcolor}{95.56}} &80.06 &\textcolor{\downcolor}{79.86} &81.63 &\textcolor{\upcolor}{84.71} &\multirow{-11}{*}{78.06} &\multirow{-11}{*}{\textcolor{\upcolor}{78.67}} \\ \midrule

\multicolumn{9}{l}{\textbf{- Training algorithms}} \vspace{.05cm} \\

G-ODIN \cite{godin20cvpr} &88.64 &\textcolor{\upcolor}{90.00} &95.94 &\textcolor{\upcolor}{96.94} &94.89 &\textcolor{\upcolor}{95.48} &72.50 &\textcolor{\upcolor}{73.83} &87.54 &\textcolor{\upcolor}{89.37} &75.10 &\textcolor{\upcolor}{77.00} \\
VOS \cite{vos22iclr} &87.44 &\textcolor{\upcolor}{89.63} &92.52 &\textcolor{\upcolor}{94.75} &95.20 &\textcolor{\upcolor}{95.62} &80.05 &\textcolor{\downcolor}{79.32} &81.12 &\textcolor{\upcolor}{84.39} &78.37 &\textcolor{\upcolor}{78.83} \\
LogitNorm \cite{wei2022mitigating} &91.48 &\textcolor{\upcolor}{91.72} &96.89 &\textcolor{\upcolor}{97.09} &94.52 &\textcolor{\upcolor}{95.27} &74.96 &\textcolor{\upcolor}{75.48} &82.60 &\textcolor{\upcolor}{85.81} &76.16 &\textcolor{\upcolor}{78.05} \\
CutMix \cite{cutmix19cvpr} &87.47 &\textcolor{\upcolor}{88.33} &92.34 &\textcolor{\upcolor}{93.18} &96.48 &96.48 &78.19 &\textcolor{\upcolor}{78.86} &76.61 &\textcolor{\upcolor}{79.18} &80.41 &\textcolor{\upcolor}{81.23} \\
CrossEntropy+ \cite{vaze2021open} &89.99 &\textcolor{\upcolor}{91.00} &93.42 &\textcolor{\upcolor}{93.81} &95.84 &\textcolor{\upcolor}{95.87} &78.92 &\textcolor{\upcolor}{79.47} &78.77 &\textcolor{\upcolor}{80.26} &77.62 &\textcolor{\upcolor}{79.29} \\ \midrule

\multicolumn{9}{l}{\textbf{- Data-driven methods}} \vspace{.05cm} \\

PixMix \cite{hendrycks2021pixmix} &92.26 &\textcolor{\upcolor}{92.94} &95.74 &\textcolor{\upcolor}{96.22} &95.47 &\textcolor{\upcolor}{96.01} &76.93 &\textcolor{\upcolor}{78.13} &84.76 &\textcolor{\upcolor}{85.99} &77.82 &\textcolor{\upcolor}{79.61} \\
OE \cite{oe18nips} &88.62 &\textcolor{\upcolor}{90.16} &97.16 &\textcolor{\downcolor}{96.92} &95.17 &\textcolor{\upcolor}{95.59} &78.00 &\textcolor{\downcolor}{77.40} &83.01 &\textcolor{\upcolor}{85.43} &77.81 &\textcolor{\upcolor}{77.99} \\ \midrule

\textbf{Average} & 86.25 &\textcolor{\upcolor}{89.04}& 92.05 &\textcolor{\upcolor}{94.11}& 95.10 &\textcolor{\upcolor}{95.64}& 76.63 &\textcolor{\upcolor}{77.20}& 80.29 &\textcolor{\upcolor}{83.65}& 77.89 &\textcolor{\upcolor}{78.74} \\

\textbf{Best} & 92.26 &\textcolor{\upcolor}{92.94}& 97.16 &\textcolor{\downcolor}{97.09}& 96.48 &96.48 & 80.22 &\textcolor{\upcolor}{80.34}& 87.54 &\textcolor{\upcolor}{89.37}& 80.41 &\textcolor{\upcolor}{81.23} \\

\midrule[0.8pt]\bottomrule[1.5pt]

\end{tabular}
}}
\vspace{-3mm}
\end{table*}

\subsection{Training with real and synthetic ID samples}
\label{sec:3.2}

Subtle differences or shifts can exist between the synthetic and real distributions \cite{cas}.
To avoid bias towards the synthetic distribution, we train the model using real and synthetic data together.
Concretely, we design a weighted objective, which is essential for obtaining superior performance as shown in later experiments:
\begin{multline}
    \min_f\left[\alpha\cdot\mathbb{E}_{(\bm{x},y)\sim \Dinr}L(\bm{x},y;f)+\right.\\ \left. \quad \quad (1-\alpha)\cdot\mathbb{E}_{(\tilde{\bm{x}},\tilde{y})\sim \Dins}L(\tilde{\bm{x}},\tilde{y};f)\right],
\label{eq:SIO_id_only_training}
\end{multline}
where $\alpha\in[0,1]$ is the weighting term.
Note that, similarly to Equation \ref{eq:id_only_training}, $L$ could be any specific loss function of existing OOD training algorithms.
SIO can also be seamlessly integrated with methods that incorporate OOD samples during the training (Equation \ref{eq:id_ood_training}).
In such cases, the objective is:
\begin{multline}
    \min_f\left[\alpha\cdot\mathbb{E}_{(\bm{x},y)\sim \Dinr,\hat{\bm{x}}\sim\Dout}L(\bm{x},y,\hat{\bm{x}};f)+\right.\\ \left. \quad \quad (1-\alpha)\cdot\mathbb{E}_{(\tilde{\bm{x}},\tilde{y})\sim \Dins,\hat{\bm{x}}\sim\Dout}L(\tilde{\bm{x}},\tilde{y},\hat{\bm{x}};f)\right].
\label{eq:SIO_id_ood_training}
\end{multline}

In practice, SIO is implemented in an equivalent but much more efficient way than its basic form in Equation \ref{eq:SIO_id_only_training} and \ref{eq:SIO_id_ood_training}.
Instead of performing two separate forward passes and computing the weighted loss, we do the weighting inside each mini-batch by replacing a certain amount of real ID samples with synthetic ones such that the ratio between real and synthetic ID samples is $\alpha:1-\alpha$ within each batch (see Figure \ref{fig:method}).
Such implementation avoids additional computation overhead and allows fair comparison to real data-only training since each model is trained under the exact same budget.

\section{Experiments}

We demonstrate the effectiveness of additional synthetic ID data for OOD detection through extensive experiments and analyses, starting with results in settings where CIFAR-10/100 \cite{cifar} is used as the ID dataset.
We then scale up to high-resolution settings with ImageNet splits.
Finally, we analyze SIO's robustness to hyperparameters.

\subsection{CIFAR}
\label{sec:4.1}

Benchmarking OOD detection methods used to be challenging because there was no unified platform with standardized implementations and benchmarks, making it difficult to make direct comparisons.
However, the recent work called OpenOOD \cite{yang2022openood} provides such a platform that enables fair and accurate benchmarking.
In this work, we use OpenOOD's setup as a basis for our experiments while making some modifications where necessary.

\noindent \textbf{Baselines.}
Since SIO is orthogonal to existing OOD detection methodologies, we demonstrate the effectiveness of SIO by combining it with multiple OOD detection approaches and comparing the results with the real data-only counterpart in each case.
Specifically, we consider 11 inference techniques \cite{openmax16cvpr,msp17iclr,guo2017calibration,odin18iclr,energyood20nips,react21neurips,species22icml,species22icml,haoqi2022vim,sun2022knnood,sun2021dice}, 5 specialized training algorithms \cite{godin20cvpr,wei2022mitigating,cutmix19cvpr,vos22iclr,vaze2021open}, and 2 data-driven methods \cite{hendrycks2021pixmix,oe18nips}, resulting in a total of 18 OOD methods.
All of them are the top-performing ones according to the OpenOOD benchmark.

\noindent \textbf{Training setup.}
We use ResNet-18 \cite{resnet} as the classifier architecture, following OpenOOD.
Regardless of the training algorithm, we train the model using Nesterov SGD with a momentum of 0.9. 
The initial learning rate is set to 0.1 and is decayed according to the cosine annealing schedule \cite{loshchilov2017sgdr}. 
A weight decay of 0.0005 is applied during training, and the batch size is set to 128. 
The only deviation from the OpenOOD default configuration is that we train each model for 200 epochs instead of 100 epochs, as longer training has been shown to improve OOD performance \cite{vaze2021open}.
For method-specific hyperparameters, we use the default or recommended values provided by OpenOOD.

For our proposed SIO, we generate 1000K synthetic ID samples using StyleGAN2 \cite{stylegan2ada} for both the CIFAR-10 and CIFAR-100 datasets.
The weighting/ratio factor $\alpha$ (Equation \ref{eq:SIO_id_only_training} and \ref{eq:SIO_id_ood_training}) is set to 0.8, \ie, there will be 20\% synthetic ID samples in each ID training batch.
We remark that the SIO model and the real data-only model undergo the exact same number of gradient steps and share the same batch size, ensuring a fair comparison.
The only difference is that the SIO model sees more diverse training samples by leveraging synthetic ID data.

\noindent \textbf{Evaluation setup.}
For CIFAR-10/100, we consider CIFAR-100/10 as near-OOD and MNIST \cite{deng2012mnist}, SVHN \cite{svhn}, Texture \cite{dtd}, and Places365 \cite{zhou2017places} as far-OOD.
Far-OOD samples are semantically far away from the ID samples and meanwhile often exhibit significant low-level, non-semantic shifts as well (due to data collection differences) \cite{ahmed2020detecting, tack2020csi}, making them easier to detect.
Near-OOD samples are more similar to ID samples in both semantic and non-semantic aspects and are more challenging to identify.

We follow the OpenOOD setting with one exception: We remove Tiny ImageNet samples from the near-OOD split since they are used as the training OOD samples for OE \cite{oe18nips}.
If not removed, the training and test OOD distributions would completely overlap, resulting in a trivial case.

We use the area under the receiver operating characteristic curve (AUROC) as the metric for evaluation.
AUROC is a threshold-independent metric for binary classification; the higher the better, and the random-guessing baseline is 50\%.
We report the detection AUROC against near- and far-OOD (averaged over the OOD sets in each split) for each method.
We also report the classification accuracy on ID test data, as a good algorithm should not trade-off ID accuracy for OOD detection performance.
All reported results are averaged over 3 independent training runs.

\begin{table*}[!ht]
\centering
\caption{Results in terms of detection AUROC (\%) on two ImageNet variants. All numbers are percentages and are averaged over 3 runs. \textbackslash{} means that the NaN error occurs when evaluating that method.}
\label{tab:imagenet}
\vspace{2mm}
\resizebox{0.95\textwidth}{!}{
{\setlength\doublerulesep{1.2pt}
\begin{tabular}{lcbcbcbcbcbcb}
\toprule[1.5pt]\midrule[0.8pt]
\multirow{3}{*}{Method} &\multicolumn{6}{c}{ImageNet-10} &\multicolumn{6}{c}{ImageNet-dogs} \\ \cmidrule(lr){2-7} \cmidrule(lr){8-13}
&\multicolumn{2}{c}{near-OOD} 
&\multicolumn{2}{c}{far-OOD}
&\multicolumn{2}{c}{ID Accuracy}
&\multicolumn{2}{c}{near-OOD} 
&\multicolumn{2}{c}{far-OOD}
&\multicolumn{2}{c}{ID Accuracy}\\ 
\cmidrule(lr){2-3} \cmidrule(lr){4-5} \cmidrule(lr){6-7} \cmidrule(lr){8-9}
\cmidrule(lr){10-11} \cmidrule(lr){12-13}
& baseline & +\method & baseline & +\method & baseline & +\method & baseline & +\method & baseline & +\method & baseline & +\method \\
\midrule\midrule

OpenMax \cite{openmax16cvpr} &84.39 &\textcolor{\upcolor}{86.19} &89.14 &\textcolor{\upcolor}{90.53} & & &95.68 &\textcolor{\upcolor}{96.01} &93.70 &\textcolor{\downcolor}{92.96} & & \\
MSP \cite{msp17iclr} &84.27 &\textcolor{\upcolor}{87.25} &88.52 &\textcolor{\upcolor}{92.03} & & &95.81 &\textcolor{\upcolor}{96.17} &94.58 &\textcolor{\upcolor}{95.80} & & \\
TempScale \cite{guo2017calibration} &84.92 &\textcolor{\upcolor}{87.75} &89.24 &\textcolor{\upcolor}{92.70} & & &97.22 &\textcolor{\upcolor}{97.37} &96.51 &\textcolor{\upcolor}{97.17} & & \\
ODIN \cite{odin18iclr} &87.30 &\textcolor{\upcolor}{89.14} &93.31 &\textcolor{\upcolor}{94.80} & & &97.19 &\textcolor{\upcolor}{97.39} &98.41 &\textcolor{\upcolor}{98.63} & & \\
EBO \cite{energyood20nips} &86.09 &\textcolor{\upcolor}{88.18} &91.74 &\textcolor{\upcolor}{93.93} & & &97.96 &\textcolor{\upcolor}{98.12} &98.51 &\textcolor{\upcolor}{98.80} & & \\
GradNorm \cite{gradnorm21neurips} &87.23 &\textcolor{\upcolor}{87.67} &95.86 &\textcolor{\downcolor}{94.71} & & &95.54 &\textcolor{\upcolor}{96.05} &98.32 &\textcolor{\upcolor}{99.27} & & \\
ReAct \cite{react21neurips} &86.21 &\textcolor{\upcolor}{87.61} &92.37 &\textcolor{\upcolor}{94.50} & & &97.98 &\textcolor{\upcolor}{98.13} &99.21 &\textcolor{\upcolor}{99.47} & & \\
MLS \cite{species22icml} &86.26 &\textcolor{\upcolor}{88.36} &91.68 &\textcolor{\upcolor}{93.92} & & &97.91 &\textcolor{\upcolor}{98.05} &98.30 &\textcolor{\upcolor}{98.62} & & \\
KLM \cite{species22icml} &77.07 &\textcolor{\upcolor}{80.52} &83.65 &\textcolor{\upcolor}{89.19} & & &\textbackslash&\textbackslash &\textbackslash&\textbackslash & & \\
VIM \cite{haoqi2022vim} &\textbackslash&\textbackslash &\textbackslash&\textbackslash & & &\textbackslash&\textbackslash &\textbackslash&\textbackslash & & \\
KNN \cite{sun2022knnood} &87.81 &\textcolor{\upcolor}{89.73} &97.79 &\textcolor{\downcolor}{97.65} & & &98.11 &\textcolor{\upcolor}{98.26} &99.84 &\textcolor{\upcolor}{99.86} & & \\
DICE \cite{sun2021dice} &87.29 &\textcolor{\upcolor}{89.81} &96.06 &\textcolor{\downcolor}{95.92} &\multirow{-12}{*}{90.00} &\multirow{-12}{*}{\textcolor{\upcolor}{91.78}} &97.26 &\textcolor{\upcolor}{97.64} &98.64 &\textcolor{\upcolor}{99.18} &\multirow{-12}{*}{75.42} &\multirow{-12}{*}{\textcolor{\downcolor}{75.34}} \\ \midrule

\textbf{Average} & 85.35 & \textcolor{\upcolor}{87.47} & 91.76 & \textcolor{\upcolor}{93.63} & 90.00 & \textcolor{\upcolor}{91.78} & 97.07	 & \textcolor{\upcolor}{97.32} & 97.60 & \textcolor{\upcolor}{97.98} & 75.42 & \textcolor{\downcolor}{75.34} \\

\textbf{Best} & 87.81 & \textcolor{\upcolor}{89.81} & 97.79 & \textcolor{\downcolor}{97.65} & 90.00 & \textcolor{\upcolor}{91.78} & 98.11 & \textcolor{\upcolor}{98.26} & 99.84 & \textcolor{\upcolor}{99.86} & 75.42 & \textcolor{\downcolor}{75.34}  \\

\midrule[0.8pt]\bottomrule[1.5pt]

\end{tabular}
}}
\vspace{-3mm}
\end{table*}

\noindent \textbf{Results.}
See Table \ref{tab:cifar}.
We start with discussing CIFAR-10 results.
The first takeaway is that SIO works well with nearly all OOD methods and helps with both near- and far-OOD detection, leading to noticeable improvements in 35 out of 36 cases (18 methods $\times$ \{near-OOD, far-OOD\}).
On average, SIO improves the near-OOD and far-OOD AUROC from 86.25\% to 89.04\% and from 92.05\% to 94.11\%, respectively.

Interestingly, we find that the performance gains brought by SIO can sometimes surpass those achieved through dedicated algorithmic design.
For example, SIO improves the near-OOD / far-OOD AUROC of the MSP detector from 86.71\% / 91.37\% to 89.86\% / 93.59\%, outperforming the more complex KNN detector on the baseline model, which achieves 89.38\% / 93.51\% at the cost of significantly increased inference latency \cite{timing}.
Furthermore, we confirm that SIO is compatible with data-driven methods that incorporate external OOD data (OE and PixMix), suggesting that the information contained in synthetic ID samples is complementary to that in external OOD samples.

Our second takeaway is that SIO improves the state-of-the-art result on the challenging CIFAR-10 near-OOD detection from 92.26\% to 92.94\%.
On CIFAR-10 far-OOD detection, SIO does not outperform vanilla OE, which incorporates a large set of external OOD samples for training.
We suspect that this is because far-OOD detection on CIFAR-10 is dominated by low-level statistics \cite{ahmed2020detecting,tack2020csi}, and using synthetic ID samples may slightly push the model away from the real low-level statistics of the ID data, causing a shrinked difference between ID and OOD samples.
Nonetheless, SIO improves far-OOD detection in all other cases and effectively closes the gap between non-OE methods and OE. 
Notably, LogitNorm + SIO yields a 97.09\% AUROC without any OOD training data, which is on par with the 97.16\% AUROC achieved by OE.

Lastly, we observe that SIO can also benefit ID classification accuracy.
While this is not the focus of this work, our finding suggests that synthetic samples can be helpful for accuracy if used properly, challenging previous beliefs \cite{cas}.
On the other hand, however, in our later experiments where we vary the hyperparameters, we find that SIO consistently boosts OOD detection performance even if it does not improve ID accuracy.
More discussion on this can be found in Section \ref{sec:analyses}.

On CIFAR-100, we observe similar results to CIFAR-10, with the general observation that SIO can benefit OOD detection performance. 
With SIO, the average near-OOD / far-OOD AUROC is lifted from 76.63\% / 80.29\% to 77.20\% / 83.65\%, and the best numbers are improved from 80.22\% / 87.54\% to 80.34\% / 89.37\%.

\subsection{Scaling to high-resolution images}

We now investigate whether SIO's effect can extend to high-resolution images at the scale of ImageNet.
However, we note that generative modeling on certain ImageNet categories remains a challenging problem due to inherent difficulties in the data \cite{biggan,cas}.
For example, the category \texttt{tench} includes many images depicting the fish being held by human beings, which causes the generative model to learn irrelevant information unrelated to the target object itself.
Consequently, generated images may appear unnatural and deviate significantly from the true distribution (see Appendix \ref{sec:ap_inet} Figure \ref{fig:inet_examples} for visual examples).
In such cases, including unrealistic synthetic images during training is unlikely to be beneficial.
This issue, however, is related to generative modeling rather than our proposed SIO and is expected to be mitigated as generative modeling techniques continue to advance. 
For our experimental purpose, here we utilize two subsets of ImageNet categories where plausible images can be generated. 
This allows us to examine the effectiveness of SIO in high-resolution settings with limited generative modeling challenges.

\begin{figure*}[t]
\centering
\begin{minipage}{0.48\textwidth}
\centering
\includegraphics[width=0.9\linewidth]{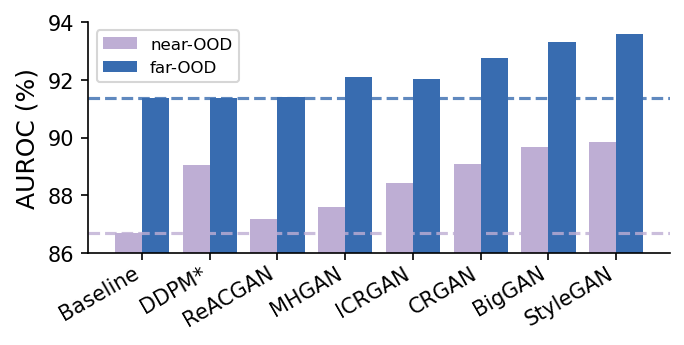}
\end{minipage}
\begin{minipage}{0.4\textwidth}
\centering
\resizebox{0.65\linewidth}{!}{
\begin{tabular}{lcc}
     data source & FID$\downarrow$
     & AUROC$\uparrow$ \\ \midrule 
     DDPM* & 3.17 & 90.22 \\
     ReACGAN & 4.40 & 89.29 \\
     MHGAN & 4.18 & 89.84 \\
     ICRGAN & 3.52 & 90.24 \\
     CRGAN & 3.48 & 90.94 \\
     BigGAN & 2.98 & 91.49 \\
     StyleGAN & 2.67 & 91.73 \\ \bottomrule
\end{tabular}
}
\end{minipage}
\vspace{-1mm}
\caption{\textbf{Left:} OOD detection results of using different generative models. SIO is fairly robust to the choice of generative model. \textbf{Right:} We find that the FID metric of synthetic images correlates well with the corresponding OOD detection performance (the average of near- and far-OOD AUROC). *Unlike GAN models, the diffusion model (DDPM) here is unconditional.}
\label{fig:gen_model}
\vspace{-3mm}
\end{figure*}

\begin{figure*}[t]
\centering
   \includegraphics[width=0.95\linewidth]{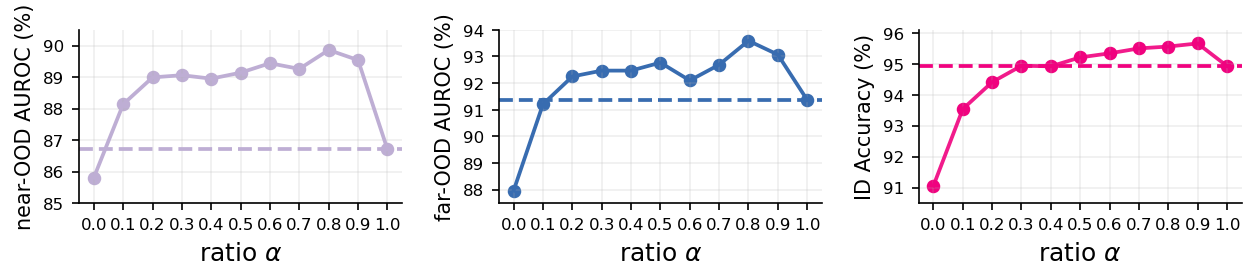}
   \vspace{-1mm}
   \caption{Results of varying the ratio $\alpha$ in Equation \ref{eq:SIO_id_only_training}. Compared with the real data-only baseline ($\alpha$=1.0), SIO consistently leads to improvements in a wide range of $\alpha$ (from 0.2 to 0.9).}
   \vspace{-1mm}
\label{fig:ratio}
\vspace{-3mm}
\end{figure*}

\noindent \textbf{Datasets.}
The first subset is ImageNet-10, which is similar to CIFAR-10 with categories such as \texttt{aircraft}, \texttt{automobile}, and \texttt{bird}.
The other one is ImageNet-dogs \cite{huang2021feature}, which consists of 100 dog categories.
See Appendix \ref{sec:ap_inet} for the complete list of class WordNet IDs.

\noindent \textbf{Training setup.}
We train ResNet-18 models for 60 epochs using SGD with a momentum of 0.9. 
The initial learning rate is 0.1 and decays according to the cosine annealing schedule. 
We use a batch size of 256 and the standard \texttt{RandomResizedCrop} with the final size being 224x224 as data augmentation.
For our SIO, we generate 10K synthetic images per category for both ImageNet-10 and ImageNet-dogs datasets using BigGAN \cite{biggan}.
The weighting/ratio factor $\alpha$ in Equation \ref{eq:SIO_id_only_training} is set to 0.9.

\noindent \textbf{Evaluation setup.}
Due to the high (sometimes unaffordable) computational cost of many specialized training methods \cite{vaze2021open,yang2022openood}, we only evaluate inference techniques with standard cross-entropy training, following OpenOOD. 
However, it should be noted that standard training is in fact a strong baseline on the ImageNet scale \cite{vaze2021open}.

For ImageNet-10, we use Species \cite{species22icml}, iNaturalist \cite{van2018inaturalist}, ImageNet-O \cite{natural_adv_examples}, and OpenImage-O \cite{haoqi2022vim} as near-OOD, as per OpenOOD.
For ImageNet-dogs, we take non-dog ImageNet samples as near-OOD, which is a more challenging problem \cite{huang2021feature}.
For both datasets, Textures \cite{dtd}, MNIST \cite{deng2012mnist}, and SVHN \cite{svhn} are used as far-OOD \cite{yang2022openood}.
In line with our previous experiments, we report the average near-OOD / far-OOD AUROC and ID accuracy from 3 runs.

\noindent \textbf{Results.}
Our results on the ImageNet splits demonstrate that the benefits of SIO observed in CIFAR experiments generalize to high-resolution images.
On ImageNet-10, while SIO does not achieve a higher best AUROC against far-OOD, it does improve the average near-OOD / far-OOD AUROC from 85.35\% / 91.76\% to 87.47\% / 93.63\% and achieves the best near-OOD AUROC of 89.81\% compared to the best score of 87.81\% from real data-only baselines.
On ImageNet-dogs, interestingly, SIO slightly degrades the ID accuracy, but still yields better average and best results than the real data-only baselines.

\subsection{Analyses}
\label{sec:analyses}

To analyze SIO's robustness to its hyperparameters, we conduct several experiments on CIFAR-10 using MSP as the detector.


\noindent \textbf{Choice of the generative model.}
In our CIFAR experiments, we used StyleGAN to generate synthetic data.
We now investigate whether SIO remains effective when using other generative models by repeating the SIO training with several other class-conditional GANs (using pre-trained models from \cite{kang2022studiogan}). 
We also consider an unconditional diffusion model \cite{ddpm}, where we use pseudo-labels for synthetic images, as discussed in Section \ref{sec:3.1}. 
The results are shown in the left panel of Figure \ref{fig:gen_model}.
We find that in all cases, SIO yields noticeable performance gains over the real data-only baseline, demonstrating that SIO is effective regardless of the specific choice of the generative model.

However, we also observe that certain generative models outperform others when used as the source of synthetic data. 
This leads us to consider metrics that can explain relative performance and guide the selection of the generative model beforehand. 
Intuitively, the \textit{quality} and \textit{diversity} are two important considerations, meaning that we want synthetic samples to be realistic and diverse such that they can provide useful information in addition to the real data.
Since advanced generative models can already provide sufficient diversity on small-scale datasets like CIFAR \cite{stylegan2ada}, we hypothesize that in our experiments the sample quality is the dominant factor.
To measure sample quality, we use the well-established Fr\'echet Inception Distance (FID) \cite{fid}, which measures the distance between the synthetic distribution and the real distribution in the Inception feature space. 
The results are shown in the right panel of Figure \ref{fig:gen_model}.
We find that the FID metric correlates well with OOD detection performance (the average of near- and far-OOD AUROC). 
For example, StyleGAN images exhibit the lowest FID (best sample quality) and lead to the highest OOD detection results. 
The only exception to this correlation is the unconditional diffusion model, where we suspect that the quality of the pseudo-labels may also have an effect.

\begin{figure}[t]
\centering
   \includegraphics[width=0.9\linewidth]{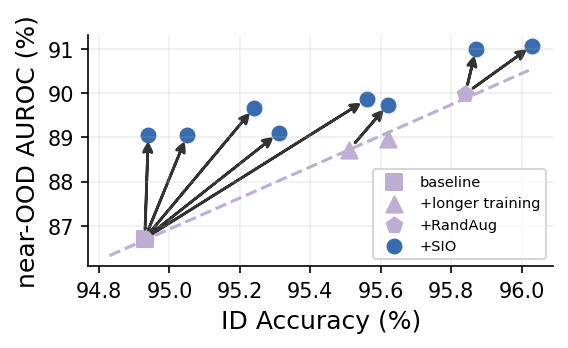}
   \vspace{-2mm}
   \caption{Near-OOD detection AUROC v.s. ID classification accuracy plot. SIO benefits OOD detection even when there are limited accuracy improvements. It also leads to a greater net gain in OOD detection per unit increase in ID accuracy, compared to the tricks (longer training and RandAug) used in \cite{vaze2021open}.}
   \vspace{-3mm}
\label{fig:oodvsid}
\end{figure}

\begin{figure}[t]
\centering
   \includegraphics[width=0.9\linewidth]{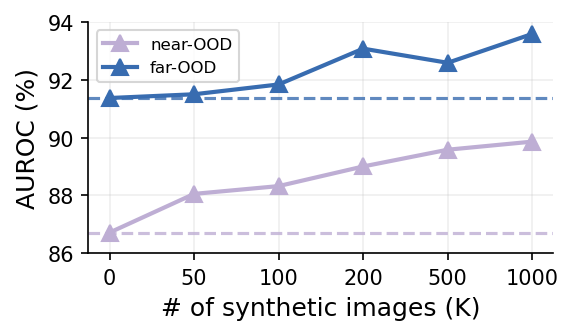}
   \vspace{-2mm}
   \caption{Results of varying the number of synthetic ID images. More samples (higher diversity) in general lead to better performance.}
   \vspace{-5mm}
\label{fig:num_syn_images}
\end{figure}

\noindent \textbf{The ratio $\alpha$.}
Another hyperparameter of SIO is the weighting/ratio $\alpha$ in Equation \ref{eq:SIO_id_only_training} and \ref{eq:SIO_id_ood_training}.
To investigate the effect of $\alpha$, we vary it from 0.0 (using only synthetic samples) to 1.0 (using only real samples) and evaluate the resulting OOD detection performance.
Results are presented in Figure \ref{fig:ratio}.

We identify that SIO consistently provides notable improvements in OOD detection performance across a wide range of $\alpha$. 
The best results are obtained under a large $\alpha$ (\eg, 0.8 or 0.9), with the majority of the samples still being real data.
We think this is because oversampling synthetic samples will exacerbate the distribution shift between the synthetic and real distribution, leading to worse performance.
Such effect can also be observed in the ID classification accuracy, where biasing towards synthetic data ($\alpha$\textless 0.5) reduces the accuracy compared with the real data baseline.
Meanwhile, notice that naively injecting synthetic images into the training set, which corresponds to $\alpha\approx$~0.05 in Figure \ref{fig:ratio} (there are 50K real samples and 1000K synthetic samples), would degrade the performance. 
Our results highlight that the weighting scheme adopted by SIO is the key ingredient of making synthetic images useful.


\noindent \textbf{Classifier architecture.}
To demonstrate that SIO can work with other classifier architectures, we repeat the CIFAR-10 experiments with DenseNet-100 \cite{densenet}.
The experimental setup and hyperparameters remain the same as before, and we focus on inference techniques for simplicity.
See Appendix \ref{sec:ap_addtional_results} for full results.
Overall, SIO improves the average scores from 84.58\% / 84.59\% to 86.04\% / 89.50\% and the best scores from 90.12\% / 93.81\% to 90.64\% / 95.43\% against near-OOD / far-OOD, respectively.

\noindent \textbf{Explaining SIO's effects.}
It is possible to assume that SIO's ability to improve OOD detection performance is solely due to its enhancement of ID classification accuracy. 
While this assumption aligns with the previous finding in \cite{vaze2021open} which suggests that there is a correlation between OOD detection performance and ID accuracy, our analysis indicates that this view does \textit{not} fully explain SIO's effects.

To demonstrate this, we reproduce the OOD-ID correlation observed in \cite{vaze2021open} by applying the techniques used in that study, including longer training and RandAug \cite{cubuk2020randaugment}.
The results of this experiment, together with the results of the SIO training, are presented in Figure \ref{fig:oodvsid}. 
The diversity of the blue dots comes from varying SIO's hyperparameters. 
We observe that SIO improves OOD detection performance even when it does not enhance ID classification accuracy. 
Additionally, we find that SIO yields a higher net gain in OOD detection performance per unit increase in ID accuracy than purely chasing a better classifier, as done in \cite{vaze2021open}.

We hypothesize that SIO's effects stem from the increased diversity provided by additional synthetic ID samples.
Our reasoning is based on the observation that neural networks often rely on ``spurious'' or ``shortcut'' features in images that are only superficially correlated with the labels \cite{yin2019fourier,ilyas2019adversarial,gilmer2019a}. 
The challenge of OOD detection, then, is that OOD samples can easily activate these spurious features. 
By training the model on more diverse ID samples, each potentially coming with different spurious features, the model may rely less on such features and become less likely to activate when presented with OOD samples.

To test our hypothesis, we vary the number of synthetic images as a proxy for the diversity of the training set.
The results presented in Figure \ref{fig:num_syn_images} provide supporting evidence for our hypothesis. 
The plot shows a generally increasing trend in OOD detection performance as the number of synthetic images increases.

\section{Conclusion}

In this work, we propose SIO, a training framework that utilizes synthetic ID samples and a weighted objective to benefit OOD detection.
SIO can be easily integrated with existing approaches and consistently improves OOD detection performance on top of multiple OOD detectors.
Importantly, our findings suggest that training with additional and diverse synthetic ID data benefits OOD detection, which may open up new directions for future research.
As generative models continue to advance, we expect SIO to remain an effective and versatile component for OOD detection.

\textbf{Public Release Number:} AFRL-2023-1371. 


{\small
\bibliographystyle{ieee_fullname}
\bibliography{egbib}
}

\appendix
\onecolumn

\section{ImageNet experiments details}
\label{sec:ap_inet}

\begin{figure*}[!h]
\centering
   \includegraphics[width=0.98\linewidth]{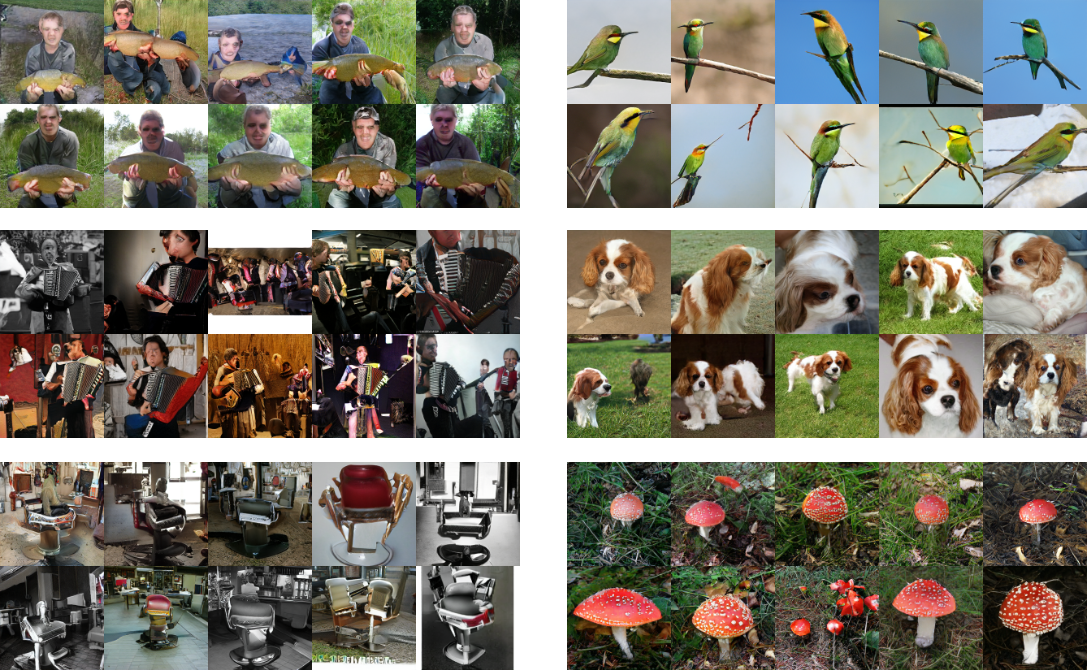}
   \caption{Examples of synthetic images generated from BigGAN on ImageNet. Certain categories (right column) are easier to model, and the synthetic images are more realistic. The model struggle to produce plausible images for other categories (left column).
   From left to right, top to bottom, the categories are n01440764: \texttt{tench}, n01828970: \texttt{bee eater}, n02672831: \texttt{accordion}, n02086646: \texttt{Blenheim spaniel}, n02791124: \texttt{barber chair}, and n12998815: \texttt{agaric}.}
\label{fig:inet_examples}
\end{figure*}

The WordNet IDs for the two ImageNet splits considered in our experiments are:
\begin{itemize}
    \item \textbf{ImageNet-10}: n02690373, n04037443, n01833805, n02123045, n01744401, n02099601, n01644373, n07745940, n04273569, n03417042.
    
    \item \textbf{ImageNet-dogs}: n02085620, n02085782, n02085936, n02086079, n02086240, n02086646, n02086910, n02087046, n02087394, n02088094, n02088238, n02088364, n02088466, n02088632, n02089078, n02089867, n02089973, n02090379, n02090622, n02090721, n02091032, n02091134, n02091244, n02091467, n02091635, n02091831, n02092002, n02092339, n02093256, n02093428, n02093647, n02093754, n02093859, n02093991, n02094114, n02094258, n02094433, n02095314, n02095570, n02095889, n02096051, n02096177, n02096294, n02096437, n02096585, n02097047, n02097130, n02097209, n02097298, n02097474, n02097658, n02098105, n02098286, n02098413, n02099267, n02099429, n02099601, n02099712, n02099849, n02100236, n02100583, n02100735, n02100877, n02101006, n02101388, n02101556, n02102040, n02102177, n02102318, n02102480, n02102973, n02104029, n02104365, n02105056, n02105162, n02105251, n02105412, n02105505, n02105641, n02105855, n02106030, n02106166, n02106382, n02106550, n02106662, n02107142, n02107312, n02107574, n02107683, n02107908, n02108000, n02108089, n02108422, n02108551, n02108915, n02109047, n02109525, n02109961, n02110063, n02110185.
\end{itemize}

\section{Additional results}
\label{sec:ap_addtional_results}

\begin{table*}[!ht]
\centering
\caption{Results in terms of detection AUROC (\%) on CIFAR-10 with DenseNet-100. All numbers are percentage and are averaged over 3 runs.}
\label{tab:cifar_dn}
{\setlength\doublerulesep{1.2pt}
\begin{tabular}{lcbcbcb}
\toprule[1.5pt]\midrule[0.8pt]
\multirow{3}{*}{Method} &\multicolumn{6}{c}{CIFAR-10} \\ \cmidrule(lr){2-7}
&\multicolumn{2}{c}{near-OOD} 
&\multicolumn{2}{c}{far-OOD}
&\multicolumn{2}{c}{ID Accuracy} \\ 
\cmidrule(lr){2-3} \cmidrule(lr){4-5} \cmidrule(lr){6-7}
& baseline & +\method & baseline & +\method & baseline & +\method \\
\midrule\midrule

OpenMax \cite{openmax16cvpr} &86.69 &\textcolor{\upcolor}{88.39} &87.29 &\textcolor{\upcolor}{91.21} & & \\
MSP \cite{msp17iclr} &88.65 &\textcolor{\upcolor}{89.36} &89.71 &\textcolor{\upcolor}{92.40} & & \\
TempScale \cite{guo2017calibration} &88.46 &\textcolor{\upcolor}{89.61} &90.34 &\textcolor{\upcolor}{93.60} & & \\
ODIN \cite{odin18iclr} &82.88 &\textcolor{\upcolor}{87.81} &83.37 &\textcolor{\upcolor}{92.40} & & \\
EBO \cite{energyood20nips} &88.91 &\textcolor{\upcolor}{90.03} &90.37 &\textcolor{\upcolor}{94.09} & & \\
ReAct \cite{react21neurips} &69.03 &\textcolor{\upcolor}{69.09} &46.76 &\textcolor{\upcolor}{59.96} & & \\
MLS \cite{species22icml} &88.89 &\textcolor{\upcolor}{90.06} &90.32 &\textcolor{\upcolor}{94.01} & & \\
KLM \cite{species22icml} &78.95 &\textcolor{\upcolor}{80.38} &81.70 &\textcolor{\upcolor}{87.60} & & \\
VIM \cite{haoqi2022vim} &82.09 &\textcolor{\upcolor}{83.43} &87.78 &\textcolor{\upcolor}{90.31} & & \\
KNN \cite{sun2022knnood} &90.12 &\textcolor{\upcolor}{90.64} &93.81 &\textcolor{\upcolor}{95.43} & & \\
DICE \cite{sun2021dice} &85.71 &\textcolor{\upcolor}{87.61} &89.06 &\textcolor{\upcolor}{93.54} &\multirow{-11}{*}{94.90} &\multirow{-11}{*}{\textcolor{\upcolor}{95.13}} \\ \midrule

\textbf{Average} & 84.58 &\textcolor{\upcolor}{86.04} & 84.59 &\textcolor{\upcolor}{89.50} & 94.90 &\textcolor{\upcolor}{95.13} \\

\textbf{Best} & 90.12 &\textcolor{\upcolor}{90.64} & 93.81 &\textcolor{\upcolor}{95.43} & 94.90 &\textcolor{\upcolor}{95.13} \\

\midrule[0.8pt]\bottomrule[1.5pt]

\end{tabular}
}
\end{table*}

\end{document}